\pdfoutput=1

\documentclass[11pt]{article}

\usepackage[]{acl}

\usepackage{pythonhighlight}
\usepackage{times}
\usepackage{listings}
\usepackage{latexsym}
\usepackage{graphicx}
\usepackage{float}
\usepackage{fancyvrb}
\usepackage{siunitx}
\usepackage[T1]{fontenc}

\usepackage[utf8]{inputenc}

\usepackage{microtype}

\title{True Detective: A Challenging Benchmark for Deep Abductive Reasoning \\in Foundation Models}

\title{True Detective: A Challenging Benchmark for Deep Abductive Reasoning}

\title{True Detective: A Challenging Benchmark for Deep Abductive Reasoning in Large Language Models}

\title{True Detective: A Deep Abductive Reasoning Benchmark \\ Impossible for GPT-3 and Challenging for GPT-4}

\title{True Detective: A Deep Abductive Reasoning Benchmark \\ Undoable for GPT-3 and Challenging for GPT-4}

\author{Maksym Del \and Mark Fishel \\
  Institute of Computer Science \\
  University of Tartu, Estonia \\
  \texttt{\{maksym,mark\}@tartunlp.ai} 
  }

\begin{document}
\maketitle
\begin{abstract}

Large language models (LLMs) have demonstrated solid zero-shot reasoning capabilities, which is reflected in their performance on the current test tasks. This calls for a more challenging benchmark requiring highly advanced reasoning ability to be solved. In this paper, we introduce such a benchmark, consisting of 191 long-form (1200 words on average) mystery narratives constructed as detective puzzles. Puzzles are sourced from the "5 Minute Mystery" platform and include a multiple-choice question for evaluation. Only 47\% of humans solve a puzzle successfully on average, while the best human solvers achieve over 80\% success rate. We show that GPT-3 models barely outperform random on this benchmark (with 28\% accuracy) while state-of-the-art GPT-4 solves only 38\% of puzzles. This indicates that there is still a significant gap in the deep reasoning abilities of LLMs and humans and highlights the need for further research in this area. Our work introduces a challenging benchmark for future studies on reasoning in language models and contributes to a better understanding of the limits of LLMs' abilities.\footnote{\url{https://github.com/TartuNLP/true-detective}}

\end{abstract}

\section{Introduction}

Large language models (LLMs) have gained significant attention in recent years due to their impressive performance on a wide range of natural language processing tasks, including reasoning tasks \cite{cot}. This calls for new, genuinely challenging benchmarks requiring LLMs to possess truly advanced reasoning capabilities to be solved. 


Abductive reasoning is a type of inference aiming at finding the minimal and most justified explanation for the set of phenomena or observations. Previous benchmarks on this topic, such as \citet{mostafazadeh-etal-2016-corpus}, consisted of short and straightforward common-sense observations and were solved by GPT models \cite{gpt}. However, the canonical example of abductive reasoning, a demanding process of a detective finding the best solution to a complex crime based on the clues and observations, was not explored as a foundation for the LLM benchmark in the literature.

Motivated by the need for a new reasoning benchmark and inspired by the complexities and particularities of a detective enterprise, we present a novel abductive reasoning benchmark consisting of 191 detective puzzles/mysteries. Mysteries are sourced from the "5 Minute Mystery" platform, where professional and aspiring authors wrote them. A puzzle is structured as a >1000 words story with 4-5 answer options. Over the last 15 years, puzzles were attempted by humans around 2000 times each with an average solve rate of 47\% (only the first try for each human for each puzzle counts). However, top human solvers (top 10) achieve a success rate of over 80\% solving more than 154 of 192 puzzles correctly. 

Moreover, additional modifications such as chain-of-thought (CoT) prompting \citet{cot, stepbystep} that are meant to invoke emergent reasoning abilities in LLMs do not help for GPT-3. 

In this study, we also assess the performance of the current state-of-the-art GPT-3 and GPT-4 models on our newly proposed dataset. We show that these models, even equipped with the Chain of Thought prompts \cite{cot, stepbystep}, are getting an accuracy rate of only 28\%, barely better than random guessing (GPT-3.5), or scoring 38\% (GPT-4), which is halfway between random guessing and average human baseline, and far behind top human solvers with their 80\% solve rate. These results reveal a significant gap in the reasoning abilities of GPT models and humans.

In our ablation study, we also supply models with golden CoTs. Golden CoTs are narratives that represent the reasoning behind the correct answer for each story (written by the mystery authors). When we attach golden CoTs to the input prompt, the best-performing GPT-3.5 model only achieves a solve rate of 63\%. This indicates LLMs' difficulty making even trivial inferences from the complex long-form story. GPT-4 models, however, get as good as the best human solvers when presented with our chain of thoughts (even though humans do not have access to the golden CoTs).

Our contributions in this paper are twofold: 
(1) a new challenging benchmark for evaluating LLMs for advanced abductive reasoning; (2) a showcase of GPT-3.5 and GPT-4 models failing to perform reasonably. 

\section{Related Work}

 \citet{mostafazadeh-etal-2016-corpus} introduced the ROCStories benchmark: narrative cloze test, which requires choosing the correct ending of the four-sentence story. \citet{bhagavatula2020abductive} expand on this dataset, requiring finding plausible explanations for narrative gaps instead of focusing on the sequence of events.
 Our benchmark contains stories of around 70 sentences that require solving the detective mystery (as opposed to simply figuring out commonsense story continuation), which is a much harder inference.

Natural language inference (NLI) is another related domain, but NLI tasks usually include much simpler and smaller inferences \cite{bowman-etal-2015-large, williams-etal-2018-broad}. \citet{swag} introduced the SWAG dataset that offers a large-scale natural language inference challenge where grounded knowledge is required to make an inference. This shares some commonality with our dataset, as some mysteries might require a share of grounded knowledge about the real world. Unlike \citet{swag}, we only offer a test set, but our stories are broader and more involved. On the other hand, \citet{biquad} introduced a question-answering benchmark that requires deeper text understanding based on the football match commentaries. Their questions range from counting the number of goals to identifying the game-winner. While answers to many of these questions are not explicitly provided in the football commentary, our mysteries require solving the whole case specifically designed to be challenging even for humans.

Lastly, \citet{cot} find that while eliciting "Chain of Thought" reasoning helps with stronger models, it can hurt when solving harder tasks with smaller models. We observe this behavior when comparing GPT-3.5 and GPT-4 on our benchmark.

\section{Benchmark}
\subsection{5 Minute Mystery Platform}
The data for this AI research was obtained from the "5 Minute Mystery"\footnote{\url{https://www.5minutemystery.com/}} online platform. This website is an online platform that has functioned for over ten years and allows users to submit and solve mysteries of varying difficulty (see Appendix \ref{sec:appendix} for an example mystery).

Based on the website author guidelines, the mysteries on the website collection are intended for readers at the sixth to eighth-grade reading level and have a recommended length of around 1200 words. To facilitate comprehension and challenge the reader, each mystery includes around four suspects and one guilty suspect. Of the 191 mysteries, the overwhelming majority ask the reader to identify the guilty suspect, with only occasional ones asking for the geographic location or the missing person. The aim is for the reader to demonstrate their abductive reasoning abilities by solving the mystery and identifying the correct solution (e.g., the murderer). Typically, one character in the story is faced with the key puzzle, and at the end of the mystery, they exclaim something like: "I figured out who is guilty!" At this point, the reader must choose the correct answer from a list of options.

In addition, mystery writers provided an explanation for the answer: a full solution (golden CoT) that elicits reasoning that leads to the correct answer. The reasoning is presented on behalf of one of the story characters (the one who says, "I know who did it" at the end of the story).  

The website also has a unique scoring system that rewards users for correctly solving mysteries, encouraging participation and engagement. In addition to providing entertainment, the website can also be used in an educational setting to help students develop their comprehension and critical thinking skills.

\subsection{Benchmark Dataset}
The mysteries in this study were obtained from the "5 Minute Mystery" (5MM) platform. We have included links to the original mysteries and to the author pages in the study, and we want to emphasize that all copyrights remain with the original authors and the 5MM team. See the authors list in the Appendix \ref{app:authors} section.

\paragraph{Dataset size and the number of answers.} The dataset used in this study consists of 191 puzzles, including 160 puzzles with four answer options, 30 puzzles with five answer options, and one puzzle with three answer options.

\begin{figure}
\centering
\includegraphics[width=0.33\textwidth]{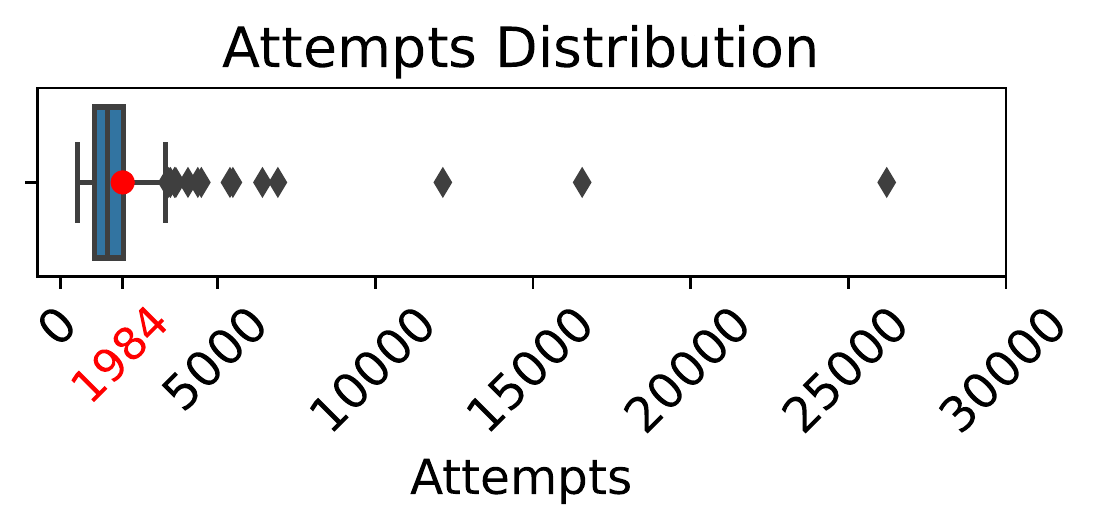}
\caption{Distribution of the number of attempts for each mystery. The red dot indicates that almost 2000 people attempted mysteries on average. This suggests that our dataset provides a robust estimate of human performance and is representative of human performance on the mysteries.}
\label{fig:attempts}
\end{figure}

\paragraph{Attempts.}
The "5 Minute Mystery" platform has been in operation for approximately 14 years and has attracted thousands of users, with over 20,000 registered by 2013. These users have made numerous attempts at each mystery, but only their first attempt is counted towards the platform's statistics. As shown in Figure \ref{fig:attempts}, the average number of attempts per mystery is 1984, with only a few puzzles being significantly more or less popular.

With such a large sample size, the resulting human performance estimate is highly robust and reliable as a benchmark.

\begin{figure}
\centering
\includegraphics[width=0.33\textwidth]{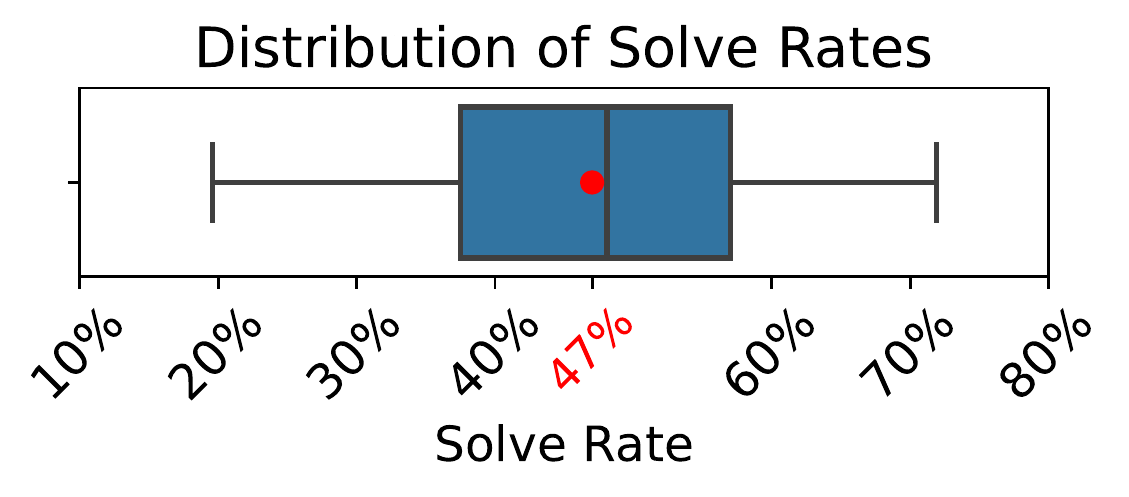}
\caption{Average human solve rate for each mystery in the dataset. The performance for most puzzles is around 40-60\%. The red dot indicates the average solve rate. This figure reveals that the majority of puzzles are challenging for human solvers, providing a good benchmark for evaluating the performance of AI models on these types of tasks.}
\label{fig:human_solve_rate}
\end{figure}

\begin{figure}
\centering
\includegraphics[width=0.35\textwidth]{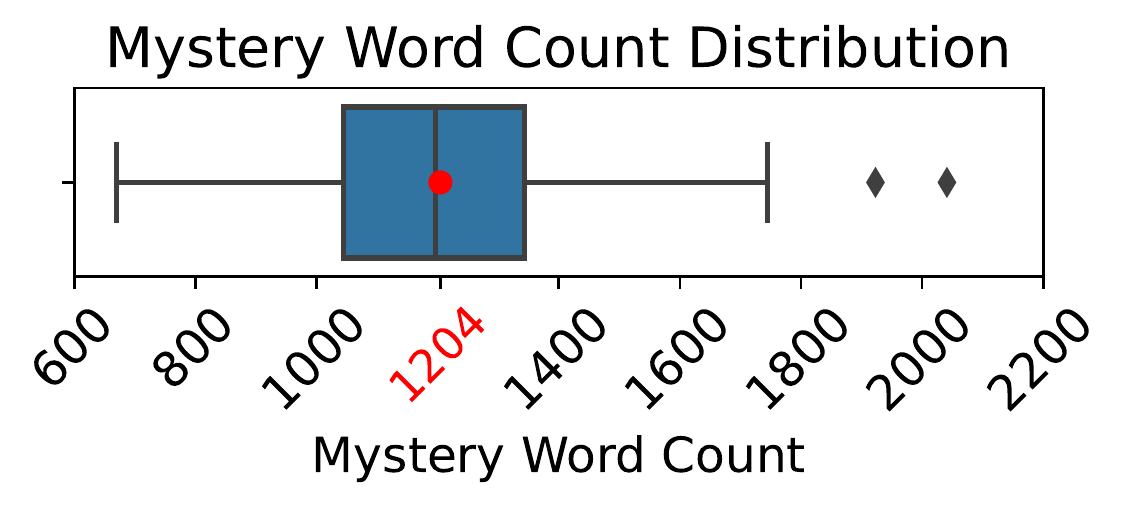}
\caption{Number of words in each mystery in the dataset. Mysteries range from 600 to around 2000 words with most of them being around 1204 words (red dot). This suggests that not only does the task require drawing highly nontrivial conclusions from the text but also doing so over relatively large texts.}
\label{fig:mystery_word_count}
\end{figure}

\begin{figure}
\centering
\includegraphics[width=0.33\textwidth]{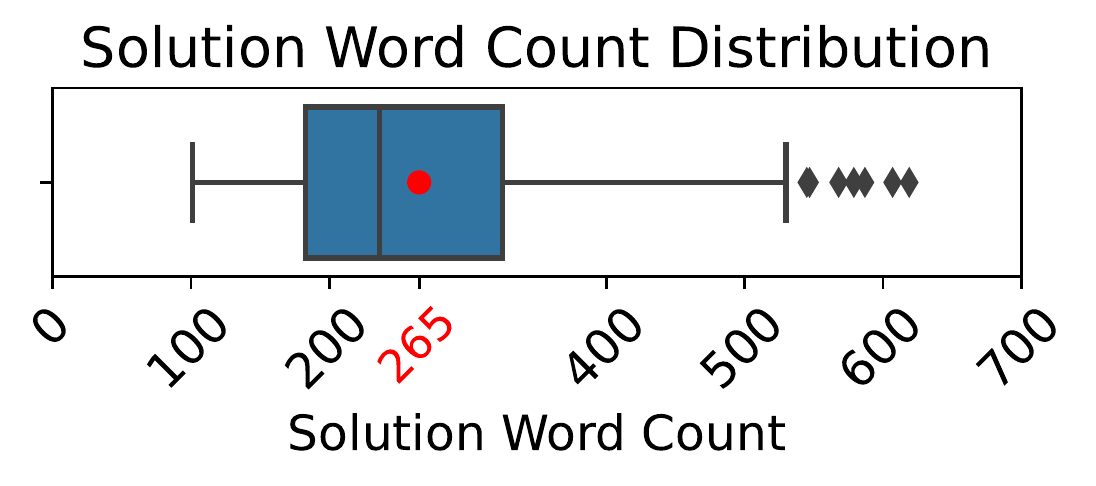}
\caption{Number of words in the solution explanations for each mystery. The red dot indicates the average number of words per explanation. This figure reveals that the average solution length is 265 words, and the longest solutions are around 600 words. Solutions (golden CoTs) are useful for a setup testing the ability of LLMs to do a trivial final answer inference over the given CoT.}
\label{fig:solution_word_count}
\end{figure}

\paragraph{Human Solve Rate.}
In the 5MM platform, human solvers have achieved moderate success. The average solve rate is 47\%, significantly higher than random guessing (around 24\%), indicating that the tasks are challenging even for humans. The top ten human solvers have an average solve rate of 80-90\%, per platform statistics. Figure \ref{fig:human_solve_rate} shows that most mysteries are solved between 40\% and 60\% of the time, with some being solved up to 70\% of the time and others close to random guessing. While the mysteries were designed to vary in difficulty, it is possible that the best explanation provided by humans may not always align with the author's intended solution for the hardest ones. However, we continue to include these mysteries in our dataset to investigate whether language models can better infer the author's intent in these cases.

\paragraph{Mystery word count.}
Figure \ref{fig:mystery_word_count} shows the distribution of the number of words in each mystery in the dataset. On average, mysteries have 1204 words, with some being as long as 2000 words. This suggests that the puzzles used in the study not only require advanced reasoning skills to solve but also require finding relevant clues from a relatively long body of text that can incriminate or exonerate suspects. This further complicates the task.

\paragraph{Golden CoTs.}
Each mystery in the dataset includes a full-text solution that provides an explanation of how one of the story characters came up with the correct answers. The average length of these solutions is 265 words, as shown in Figure \ref{fig:solution_word_count}. The solution lengths do not vary significantly, with the longest solution being around 600 words.

These solutions can be considered as ground-truth Chains-of-Thought (cite paper here), which provide insight into the author's reasoning for each puzzle. This information is valuable for a few reasons. First, it can be used as part of few-shot learning examples (again, cite). Second, as we demonstrate in Section \ref{sec:eval}, we can use these Chains-of-Thought to simplify the abductive reasoning task and evaluate whether language models can perform inference when the solution is strongly hinted at.

\section{Evaluation}
\label{sec:eval}

\subsection{Models}
The models used in this study are the InstructGPT-3.5 models \textit{GPT-3.5 (FeedME)}, \textit{GPT-3.5 (PPO)} \cite{OpenAI}, and \textit{GPT-4} \cite{gpt4}. They are causal language models based on the Transformer architecture \cite{transformer} featuring supposedly around 175B parameters for GPT-3.5s. 

\paragraph{GPT-3.5 (FeedME):} a model was trained using the FeedME method, a supervised fine-tuning method based on human-written instructions and model samples \cite{Ouyang2022TrainingLM, OpenAI}.

\paragraph{GPT-3.5 (PPO):} is a more performant update over \textit{GPT-3.5 (FeedME)} model. Apart from instruction tuning, it was also calibrated with RLHF, a reinforcement learning method that uses reward models trained from human comparisons \cite{NEURIPS2020_1f89885d, OpenAI}.

\paragraph{GPT-4:} state-of-the-art commercial model from OpenAI. Achieves human parity on multiple extremely challenging tasks \cite{gpt4}.

\subsection{Methods}
In this study, we tested GPTs in a zero-shot manner in three scenarios. This subsection outlines them.

\paragraph{Vanilla:} This method involves the task description, mystery body, and an immediate request for the final answer \cite{NEURIPS2020_1457c0d6}.

\paragraph{CoT:} This method asks LLMs to generate a Chain-of-Thought first \cite{cot, stepbystep} and only then requests the final answer. Chain-of-thought, if reasonable, allows the model to approach complex problems gradually and  unlocks strong reasoning abilities at a particular model scale \cite{cot}.   

\paragraph{Golden CoT:} This method involves generating answers to instruction-based questions by using a set of ground-truth Chain-of-Though solutions included as part of the prompt. 
This significantly simplifies the task for the model as it does not need to come up with CoT, so we can test how much of the performance depends on the CoT and how much on the final abductive reasoning step.

\subsection{Prompt Templates}
\begin{figure}[H]
\centering
\includegraphics[width=0.33\textwidth]{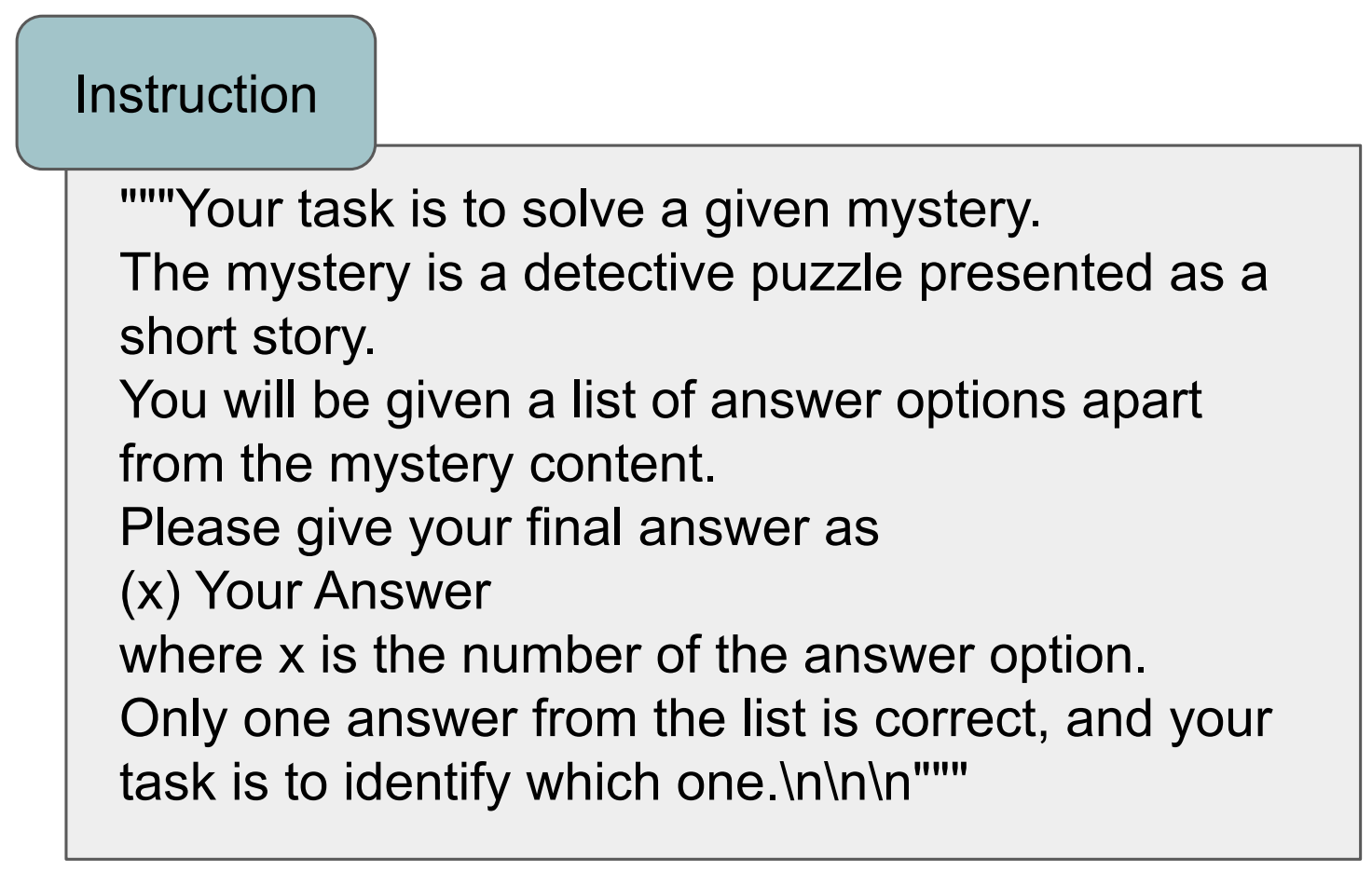}
\caption{Task instruction that we use as a prompt prefix.}
\label{fig:instuct}
\end{figure}

Figure \ref{fig:instuct} shows the task instruction that we give to the InstructGPT models at the beginning of the prompt. 

Then we always add the mystery name, list of suspects, and mystery content (body) to the prompt.





When we want to invoke Chain-of-Though reasoning, we also append the following:

\begin{small}
\begin{verbatim}
Full answer: 
Let's think step by step.
\end{verbatim}
\end{small}

When we want to provide a golden Chain-of-thought, we append the following prompt:

\begin{small}
\begin{verbatim}
Solution: 
{solution}
\end{verbatim}
\end{small}

Finally, we always ask for the final answer with

\begin{small}
\begin{verbatim}
Final answer:
\end{verbatim}
\end{small}

\subsection{Results and Discussion}
The evaluation results shown in Table \ref{tab:eval} indicate that the performance of both \textit{davinci} models under both \textit{Vanilla} and \textit{CoT} prompting scenarios is close to random. In our analysis, we also found that there is no correlation between the length of the mystery or human solve rate and the GPT's correctness.



Our Golden CoT ablation study (Table \ref{tab:eval}) demonstrates that even with relevant explanatory CoT, GPT-3.5s can only solve 63\% of puzzles correctly, suggesting that difficulty lies not only in generating the correct theory for the crime but also in making final inferences when all information is available. On the other hand, GPT-4 does not help such a problem with 83\%. 

CoT performance of GPT-3.5 models show small to no gains in performance compared to Vanilla. As indicated in \citet{cot}, a similar decrement (between GPT-3 and smaller models) was observed in models that weren't sufficiently powerful for the task suggesting that the GPT-3.5 models might also not be strong enough to generate CoT chains that would benefit the task. On the other hand, CoT GPT-4 performs better, although still underachieving compared to the average human solve rate.

The complexity of the long-form multi-character narrative and the level of reasoning required to solve the detective puzzle makes our benchmark especially difficult and sets it apart.

\begin{table}[]
\small
\centering
\begin{tabular}{lc}
\hline
\textbf{Method} & \textbf{Solve rate}\\
\hline
Random guess & 0.24  \\
Human average & 0.47  \\
Human top & 0.8-0.9 \\
\hline
Vanilla \\
\quad GPT-3.5 (FeedME) & 0.28\\
\quad GPT-3.5 (PPO) & 0.26\\
\quad GPT-4 & 0.27\\
CoT \\
\quad GPT-3.5 (FeedME) & 0.26\\
\quad GPT-3.5 (PPO) & 0.29\\
\quad GPT-4 & \textbf{0.38}\\
\hline
Golden CoT* \\
\quad GPT-3.5 (FeedME) & 0.46\\
\quad GPT-3.5 (PPO) & 0.63\\
\quad GPT-4 & 0.83\\
\hline

\end{tabular}
\caption{\label{tab:eval} Performance of GPT-3.5 (FeedME), GPT-3.5 (PPO), and GPT-4 under different prompting scenarios against the human baseline. Both vanilla task formation (Instruction and immediate answer request) and "step-by-step" chain-of-thought approaches perform almost equivalent to random guess. Even in unfair comparison, GPTs cannot match/outperform top human solvers when provided with golden chains of thought.   
}
\end{table}

\begin{figure}
\centering
\includegraphics[width=0.45\textwidth]{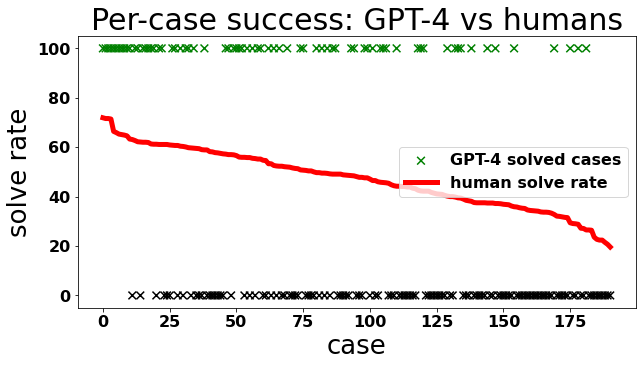}
\caption{Red line indicates case difficulty for humans, green points indicate cases where GPT-4 (CoT) solved the case successfully, and black points are for cases where GPT-4 failed. Black points are crowded on the right and green points are crowded on the left which correlates with hard and easy cases (as per humans) respectively. Therefore, GPT-4 and humans find similar cases easy/difficult.}
\label{fig:gpt4}
\end{figure}

Finally, we explore the complexity of the cases that GPT-4 (CoT) found easier or harder to manage. In our study, we did not observe a direct correlation between the length of a mystery and the level of difficulty it presented. However, when considering the level of concurrence between human decisions and those made by GPT-4, Figure \ref{fig:gpt4} demonstrates a considerable degree of agreement. Specifically, the cases perceived as challenging or straightforward by the GPT-4 were often viewed similarly by human subjects.

\section{Conclusion}

We presented a new benchmark in the form of detective puzzles to evaluate the abductive reasoning capabilities of Large Language Models. Results from state-of-the-art GPT-3.5 models across three prompting strategies showed poor performance close to random. GPT-4 managed to show comparably solid performance (when prompted with CoT), but even this model is behind the average human solve rate on our benchmark. When provided with golden CoTs, which significantly simplifies the task, GPT-4 shows good performance, while GPT-3 is still unable to do a final inference well enough. Overall, our benchmark offers insights into LLMs' limitations and provides a difficult challenge for future research on abductive reasoning in large LMs.

\section{Limitations}
Our evaluation focused solely on the performance of leading-edge GPT models details and weights of which are not publicly available. However, there is potential value in extending this study to incorporate other models like PaLM (\citeauthor{palm}) or LLaMA (\citeauthor{llama}), which we have earmarked for future research. 

Also, as the performance for average humans is only 47\% it is possible that some mysteries are ill-defined or unreasonably complicated. Among the top 10 human solvers, the solve rate is also only around 80-90\%, and GPT-4 only solves 83\% of tasks when provided with ground truth CoTs which drastically simplifies the task.


\bibliography{anthology,custom}
\bibliographystyle{acl_natbib}

\appendix

\section{Example: The Easter Egg Mystery}
\label{sec:appendix}
This appendix provides the most attempted mystery under 700 words as an example. Copyright belongs to the mystery author.

\paragraph{Metadata}
\begin{itemize}
    \item Mystery Name: The Easter Egg Mystery
    \item Author: Tom Fowler\footnote{\url{https://www.5minutemystery.com/author/tfowler}}
    \item Solve Rate: 60.8\%
    \item Attempts: 1871
    \item Answer options: (a) Anna; (b) Cole; (c) Justin; (d) Lizzie; (e) Rachel.
\end{itemize}

\paragraph{Mystery Body}
Karen Sheldon had loved Easter egg hunts ever since she was a little girl. That is why she eagerly volunteered to assist with this year’s Hunt for the children at her church.

This year, the Children’s Day Out mothers decided to do something different. Because there were so many children of all ages in the congregation, they split the hunt up into age groups. Karen’s job was to oversee several of the 6-10 year olds.

Within her group were five children she knew well. They were Rachel Smithson, whose mother Karla had volunteered to help a very grateful Karen, Justin Bates, a classmate of Rachel’s, Karen’s daughter Lizzie, Lizzie’s best friend Anna Laughlin and Cole Bryant, who was also the Sheldon’s next door neighbor.

The Easter egg hunt was on Saturday morning, the day before Easter Sunday. It was held in the large field in back of the church. Karen and Karla were grateful that today was sunny and warm although it was a bit windy. Karen was excited as the children prepared for the hunt, which was to begin at 10:00 am and last for one hour. Just before the start whistle blew, Karen told the children, “I have placed a golden Easter egg in our hunting area. There is an extra bag of candy for the child who finds it.” Only Karla and she knew that the golden egg was placed in back of the largest tree in the field, an old oak in the far corner to the left of where she and the children now stood and an area dedicated to the 6-10 year old age group.

During the hunt, Karen and Karla visited while they watched the egg hunt. During the hunt, Karen noticed that Cole stayed focused on the evergreen shrubbery in the middle of the field, finding several eggs there, much to his delight.

Karen was amused when Rachel ran to her mother and told her, “I have found a lot of eggs. I’m heading back to the rock pile. I bet I will find the golden egg there!” The rock pile was to the right of the evergreen shrubbery.

In the middle of the hunt, Karen excused herself to go inside the church to get a drink of water and sit for a few minutes. When she returned, Karla told her, “I had to run over and warn Lizzie to be careful of the dead branches on the big oak tree. One of them fell last week, hitting one of the older kids.”

As the hunt began to wind down, Karla walked out to speak with a very agitated Anna. After returning to Karen, she told her, “Anna is upset because she has found only a few eggs. I told her to keep looking; there are still a few minutes to go.” Karen noticed that Anna stayed close to Karla for the remainder of the hunt.

As the whistle blew to end the hunt, Karen walked to the center of the field to wave Justin back in. He was in the far right corner of the field, where he had been for the entire hunt. There was a sand pit in that area and Justin found several eggs there.

As the kids headed back to the start area, Karen once again excused herself to go inside. The wind had blown a speck of dust in her eye when waving Justin down and it was very painful. When she returned from rinsing her eyes, Karla and the five children were smiling at her. She asked, “What’s up?”

Karla answered, “One of our kids found the golden egg. We want you to guess which one.”

Karen smiled in return, saying, “So that’s it!” Thinking for a moment, she said, “I only have one question. When I was inside the first time, did any of the children move from one side of the field to another?”

Karla answered, “No.”

Karen tousled Justin’s hair and said, “Good. Then I know who has the golden egg!”

\subsection{Golden CoT and Answer}
\paragraph{Golden CoT.} "Good naturedly, Karla exclaimed, “How do you know?”
Smiling at Anna, she answered, “It’s not too hard to figure out. Let me explain.” The eyes of all of the children and Karla were upon her as she continued, “I placed the golden egg behind the big oak tree.” Smiling next at Cole, she said, “Cole spent the entire hour in the shrubbery, in the middle of the field, far away from the oak tree.” She patted Rachel’s shoulder and said; “Rachel did all of her hunting in the rock pile, even farther away from the oak tree.” Looking back at Anna, Karen said, “I know you don’t have the golden egg, sweetie. You were upset that you had so few eggs with only a few minutes left in the hunt and stayed close to Karla until the whistle blew.” Patting her hand, she added, “I’m sure you will do better next year.” Turning to Justin, Karen said, “You were farther away from the oak than anyone. You spent the whole hour far out in the sand pit. I even had to come get you because you could not hear the whistle.”
All eyes turned toward Lizzie. Her mother said, “So, you must have the egg. Karla told me she had to warn you of the dead branches on the oak. You were the only one near it.” Pausing, she added, “I hope everyone believes that I did not tell you where I put that egg!
Karla jumped in, “Of course we do not think that!” All of the kids echoed their support.
Lizzie broke the silence. She said, “I didn’t know about the egg until Mother told everyone else before the hunt.” Walking over to her side, Lizzie looked at Anna and offered her the golden egg, saying, “I would like for you to have this.” Tearfully, Anna thanked her friend, saying, “This is the best Easter egg hunt ever!”
Karen was so proud of Lizzie that she heartily agreed with Anna."
\paragraph{Answer:} (d) Lizzie

\section{Mystery Authors Acknowledgment}
\label{app:authors}
We thank all the 5MM mystery writers: Moe Zilla, Tom Fowler, William Shepard, Laird Long, Robbie Cutler, Barney Parmington, Stefanina Hill, Steve Shrott, Nick Andreychuk, Nicholas LeVack, Ernest Capraro, Andrea Hein, Doug Fellin, Tammy-Lee Miller, Meghan Ford, Brad Marsh, Susanne Shaphren, Randy Godwin, Ryan Hogan, Matthew Lieff, Perry McCarney, Nicholas Lovell, Mike Wever, Meg A.  Write, Elsa Darcy, PIP Writer, Julie Hockenberry.

\end{document}